\documentclass{article}

\usepackage[preprint]{nips_2018}

\usepackage[utf8]{inputenc} 
\usepackage[T1]{fontenc}    
\usepackage{hyperref}       
\usepackage{url}            
\usepackage{booktabs}       
\usepackage{amsfonts}       
\usepackage{nicefrac}       
\usepackage{microtype}      
\usepackage{float}
\usepackage{graphicx}
\usepackage{wrapfig}
\usepackage{amsmath}
\usepackage{todonotes}
\usepackage[toc,page]{appendix}
\usepackage{subcaption,graphicx}

\usepackage{xcolor}
\hypersetup{
    colorlinks,
    linkcolor={blue!50!black},
    citecolor={blue!50!black},
    urlcolor={blue!50!black}
}

\newcommand{\rulesep}{\unskip\ \vrule \hspace{1mm}}

\usepackage{authblk}

\title{High-Resolution Mammogram Synthesis\\using Progressive Generative Adversarial Networks}

\author[$\;$,$\dag$]{Dimitrios Korkinof\thanks{Corresponding author: d.korkinof10@alumni.imperial.ac.uk}}
\author[$\dag$]{Tobias Rijken}
\author[$\dag$]{Michael O'Neill}
\author[$\dag$]{Joseph Yearsley}
\author[$\dag$]{Hugh Harvey}
\author[$\dag$,$\S$]{Ben Glocker}
\affil[$\dag$]{Kheiron Medical Technologies Ltd.}
\affil[$\S$]{Department of Computing, Imperial College London}

\date{\today}

\begin{document}

\maketitle

\begin{abstract}
The ability to generate synthetic medical images is useful for data augmentation, domain transfer, and out-of-distribution detection. However, generating realistic, high-resolution medical images is challenging, particularly for Full Field Digital Mammograms (FFDM), due to the textural heterogeneity, fine structural details and specific tissue properties. In this paper, we explore the use of progressively trained generative adversarial networks (GANs) to synthesize mammograms, overcoming the underlying instabilities when training such adversarial models. This work is the first to show that generation of realistic synthetic medical images is feasible at up to 1280x1024 pixels, the highest resolution achieved for medical image synthesis, enabling visualizations within standard mammographic hanging protocols. We hope this work can serve as a useful guide and facilitate further research on GANs in the medical imaging domain.
\end{abstract}

\section{Introduction}

The generation of synthetic medical images is of increasing interest to both image analysis and machine learning communities for several reasons. First, synthetic images can be used to improve methods for downstream detection and classification tasks, by generation of images from a particularly sparse class, or by transforming existing images in a plausible way to generate more diverse datasets (known as data augmentation). \cite{xrayGAN} and \cite{frid2018synthetic} show the benefits of this approach as applied to chest X-ray and liver lesion classification, respectively. \cite{Costa2018} successfully use generative adversarial networks (GANs) in an image-to-image translation setting to learn a mapping from binary vessel trees to realistic retinal images.

Second, GANs can be used in domain adaptation, in which a model trained on images of one domain is applied to images of another domain where labels are scarce or non-existent. Images across related modalities can have significantly different visual appearance, such as in the cases of CT and MRI, or across different hardware vendors or even when using different imaging protocols. As a result, transferring a model across domains can severely degrade its performance. To that end, \cite{kamnitsas2017unsupervised} used adversarial training to increase the robustness of segmentation in brain MRI and \cite{lafarge2017domain} in histopathology images.

Third, image-to-image translation using GANs has achieved impressive results in several applications, such as image enhancement (i.e. denoising \citep{low_dose_CT_2}, super-resolution \citep{ledig2017superres} etc) and artistic style transfer (i.e. \citep{ulyanov2017texture}). Especially the former, has been shown to be successful in enhancing images from low-dose CT scans so that they become comparable with high-dose CTs, as shown in \cite{wolterink2017generative} and \cite{low_dose_CT_2}.

Finally, in semi-supervised learning, an adversarial objective can help to leverage unlabeled alongside labeled data in order to improve classification or detection performance. We refer to \cite{lahiri2017generative} for an example of semi-supervised learning as applied to retinal images.

In recent years GANs have lead to breakthroughs in a number of different non-medical applications involving generation of synthetic images, such as single-image super-resolution \citep{ledig2017superres}, image-to-image translation \citep{isola2016} and the generation of artistic images \citep{elgammal2017} to name a few.

GANs manage to ameliorate many of the issues associated with other generative models. For instance, auto-regressive models \citep{van2016pixel} generate image pixels one at a time, conditioned on all previously generated pixels, by means of a Recurrent Neural Network (RNN). These methods have shown promise, however have not yet been able to scale to high image resolutions. Additionally, the computational cost of generating a single image does not scale favorably with its resolution. With Variational Auto-encoders (VAEs) \citep{kingma2013}, restrictions on the prior and posterior distributions limit the quality of the drawn samples. Furthermore, training with pixel losses exhibits an averaging effect across multiple possible solutions in pixel space, which manifests itself as blurriness (discussed in more detail in \cite{ledig2017superres}). In contrast, GANs are able to produce samples in a single shot and do not impose restrictions on the generating distribution in a process similar to sampling from the multitude of possible solutions in pixel space, which generally leads to sharper and higher quality samples.

\begin{figure}[t!]
  \begin{center}
    \includegraphics[width=0.8\textwidth]{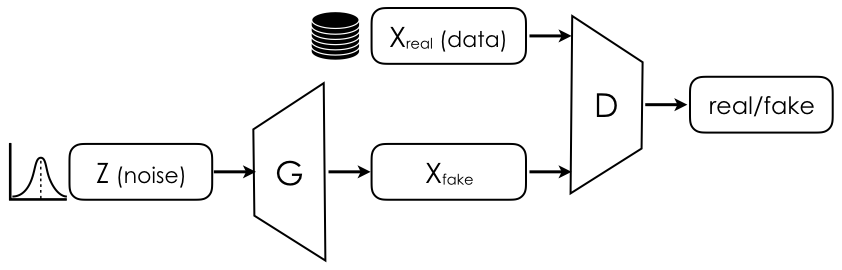}
    \caption{Schematic representation of a generative adversarial network.} \label{fig:GANs}
    \end{center}
\end{figure}

The framework for training generative models in an adversarial manner was first introduced in the seminal work of \cite{goodfellow2014}. This framework is based on a simple but powerful idea: the generator neural network aims to produce realistic examples able to deceive the discriminator which aims to discern between original and generated ones (a `critic'). The two networks form an adversarial relationship and gradually improve one-another through competition, much like two opponents in a zero-sum game (see Fig. \ref{fig:GANs}). The main disadvantage is that training these models requires reaching a Nash equilibrium, a more challenging task than simply optimizing an objective function. As a result, training can be unstable, susceptible to mode collapse and gradient saturations \citep{WassersteinGAN}.

Stabilizing GAN training becomes even more pertinent as our aim shifts towards high resolution images, such as medical images, where the dimensionality of the underlying true distribution in pixel space can be enormous and directly learning it may be unattainable. A key insight made in \cite{pggan} is that it is beneficial to start training at a low resolution, before gradually increasing it as more layers are phased in. This was shown not only to increase training stability at high resolutions, but also to speed up training, since, for much of the training, smaller network sizes are used.

The goal of this paper is to demonstrate the applicability of GANs in generating synthetic FFDMs. Mammograms contain both contextual information indicative of the breast anatomy and a great level of fine detail indicative of the parenchymal pattern. The large amount of high frequency information makes it imperative for radiologists to view these images in high-resolution. For instance, the spiculation of a mass or certain micro-calcification patterns as small as 1-2 pixels in diameter can indicate malignancy and are thus very important to consider. Our aim was to train a generator convolutional neural network (CNN) able to produce realistic, high-resolution mammographic images. For that purpose we attempted to follow \cite{pggan} as closely as possible, as our previous attempts generating even low-resolution images did not yield acceptable results.

The rest of the paper is arranged as follows: In Section \ref{GAN_theory} we summarize the key theoretical underpinnings of GANs, including their progressive training, and various stabilization methods that can be employed. In Section \ref{proposed_approach} we outline our methodology that builds on previously published literature and discuss the results of our experiments in detail. Finally, in the Appendix, readers can find several visual examples of successes and failures of the developed approach.

\section{Generative Adversarial Networks}\label{GAN_theory}

\subsection{Theoretical Background}

The framework for training generative models in an adversarial manner consists of a \textit{generator} $G$, tasked with generating samples highly probable under the true data distribution, and a \textit{discriminator} $D$, tasked with distinguishing synthesized samples from original ones. Both the generator and discriminator are trained using cost functions directly opposing each other, which can be regarded as either a zero-sum game or as a saddle point optimization problem.

The original GAN value function \citep{goodfellow2014} is expressed in terms of the discriminator binary cross entropy, as follows:

\begin{equation}
\underset{G}{\min}\underset{D}{\max}\left[\underset{\boldsymbol{x}\sim\mathbb{P}_{\boldsymbol{x}}}{\mathbb{E}}\left[\log D\left(\boldsymbol{x}\right)\right]+\underset{\boldsymbol{z}\sim\mathbb{P}_{\boldsymbol{z}}}{\mathbb{E}}\left[\log\left(1-D\left(G\left(\boldsymbol{z}\right)\right)\right)\right]\right]
\label{eq:GAN_obj}
\end{equation}

where as $\mathbb{P}_{\boldsymbol{x}}$ we denote the true data distribution with $\boldsymbol{x}$ being a sample real image and as $\mathbb{P}_{\boldsymbol{z}}$ we denote a proxy random distribution of our choice (usually uniform) with $\boldsymbol{z}$ being a random latent vector drawn from it. Note that the distributions we aim to match are $\mathbb{P}_{\boldsymbol{x}}$ and $\mathbb{P}_{\tilde{\boldsymbol{x}}}$, with $\tilde{\boldsymbol{x}}=G\left(\boldsymbol{z}\right)$ being a randomly generated image. 

The above objective was shown to be equivalent to minimizing the Jensen-Shannon divergence between the generator and true data distributions \citep{goodfellow2014}.

Training is typically performed using batch discrimination \citep{Salimans2016}, according to which the networks are presented with batches sampled exclusively from the true or generated images at each time-step.

An issue that commonly arises due to the objective in Eq. \ref{eq:GAN_obj} is that, if the discriminator becomes too effective at discerning real from generated images, the second term of the objective approaches zero and the gradient to the generator vanishes. As a result the generator cannot further improve and training ceases. In order to avoid this undesired behaviour, the authors in \cite{goodfellow2014} suggest to alternatively minimize $-\log\left(D\left(G\left(\boldsymbol{z}\right)\right)\right)$ with respect to the generator parameters, which is directly equivalent, but does not enjoy the same theoretical justification.

\subsection{Wasserstein GAN}

In \cite{WassersteinGAN}, the authors show how the original GAN objective in Eq. (\ref{eq:GAN_obj}) is potentially discontinuous with respect to the generator's parameters, which leads to instability during training. They proposed a new objective based on the Wasserstein distance (a.k.a. the earth mover's distance) to remedy this. Intuitively, the Wasserstein distance $W(p, q)$ is the minimum cost of transporting probability mass in order to transform one distribution $p$ into another distribution $q$, where the cost is the mass multiplied by the transport distance. Under mild assumptions, $W(p, q)$ is continuous everywhere and differentiable almost everywhere, which the authors claim leads to improved stability during optimization.

Formally, the Wasserstein GAN objective function is defined using the Kantorovich-Rubinstein duality \citep{OptTrans} as: 

\begin{equation}
\underset{\theta}{\text{min}}\underset{\boldsymbol{w} \in \mathcal{W}}{\text{max}}
\left[
\underset{\boldsymbol{z}\sim\mathbb{P}_{\boldsymbol{z}}}{\mathbb{E}}\left[f_{\boldsymbol{w}}\left(g_{\boldsymbol{\theta}}\left(\boldsymbol{z}\right)\right)\right]-\underset{\boldsymbol{x}\sim\mathbb{P}_{\boldsymbol{x}}}{\mathbb{E}}\left[f_{\boldsymbol{w}}\left(\boldsymbol{x}\right)\right]
\right]
\end{equation}

where $f_{\boldsymbol{w}}\left(\cdot\right)$ is the critic function transforming an image to a discriminative latent feature space, as opposed to previously being trained to discern between original and generated images, and $\{f_{\boldsymbol{w}}\}_{\boldsymbol{w} \in \mathcal{W}}$ is the set of all critic functions that are $1$-Lipschitz continuous.

To enforce the Lipschitz continuity on the critic it is sufficient to clip the weights $\boldsymbol{w}$ of the critic to lie within a compact space $[-c, c]$ \citep{WassersteinGAN}. However, as \cite{ImprovedWassersteinGAN} show, this clipping can lead to optimization problems. Instead, they propose adding a gradient penalty term to the Wasserstein objective as an alternative way to ensure the Lipschitz constraint. Their improved Wasserstein objective used in this work, is formulated as follows:

\begin{equation}
\mathcal{L}=\underset{\boldsymbol{z}\sim\mathbb{P}_{\boldsymbol{z}}}{\mathbb{E}}\left[f_{\boldsymbol{w}}\left(g_{\boldsymbol{\theta}}\left(\boldsymbol{z}\right)\right)\right]-\underset{\boldsymbol{x}\sim\mathbb{P}_{\boldsymbol{x}}}{\mathbb{E}}\left[f_{\boldsymbol{w}}\left(\boldsymbol{x}\right)\right]+\lambda\underset{\hat{\boldsymbol{x}}\sim\mathbb{P}_{\hat{\boldsymbol{x}}}}{\mathbb{E}}\left[\left(\left\Vert \nabla_{\hat{\boldsymbol{x}}}f_{\boldsymbol{w}}\left(\hat{\boldsymbol{x}}\right)\right\Vert _{2}-\beta\right)^{2}\right]
\label{eq:wggp}
\end{equation}

where $\hat{\boldsymbol{x}}$ is a random interpolation between an original and a generated image, $\hat{\boldsymbol{x}}=\gamma\boldsymbol{x}+(1-\gamma)g_{\boldsymbol{\theta}}\left(\boldsymbol{z}\right),\:\gamma\sim\mathcal{U}(0,1)$ and the hyper-parameter $\beta$ is the target value of the gradient magnitudes, usually selected $\beta$=1.

\subsection{Stabilization Methods}

Despite their improved stability, even Wasserstein GANs remain notoriously difficult to train and subject to instabilities when the equilibrium between the generator and discriminator is lost. The problem stems from the fact that the optimal point of the joint GAN objective corresponds to a saddle point, which alternating SGD methods such as those used to train the generator and discriminator networks do not reliably converge to. 

A lot of research is being dedicated to stabilizing convergence to this saddle point. \cite{yadav2018stabilizing} combine SGD with a `prediction step' that prevents `sliding off' the saddle due to maximization with respect to the discriminator overpowering minimization of the generator or vice-versa. \cite{Adolphs2018} exploit curvature information to escape from undesired stationary points and converge more consistently to a desired optimum. Finally, \cite{daskalakis2018} use Optimistic Mirror Descent (OMD) to address the limit oscillatory behavior known to impede saddle point convergence.

Stable GAN convergence becomes even more elusive when high resolution images are involved. In this setting differences between the high frequency artifacts of original and generated images make it even easier for the discriminator to win out over the generator, destabilizing training. Progressively trained GANs, which we describe next, were developed to tackle this problem.

\begin{figure}[t!]
  \begin{subfigure}[t]{0.32\textwidth}
    \includegraphics[width=\textwidth]{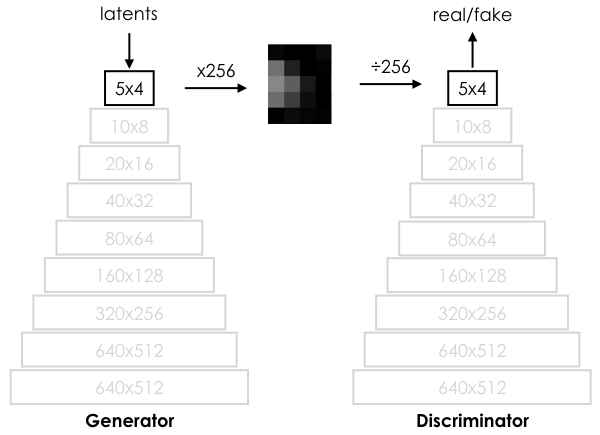}
    \caption{Scale 0}
  \end{subfigure}
  \rulesep
  \begin{subfigure}[t]{0.32\textwidth}
    \includegraphics[width=\textwidth]{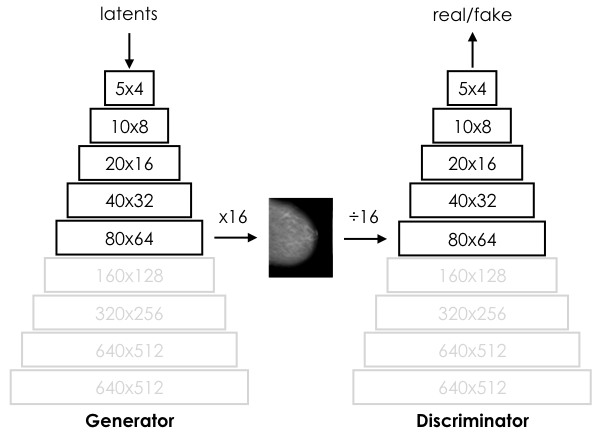}
    \caption{Scale 4}
  \end{subfigure}
  \rulesep
  \begin{subfigure}[t]{0.32\textwidth}
    \includegraphics[width=\textwidth]{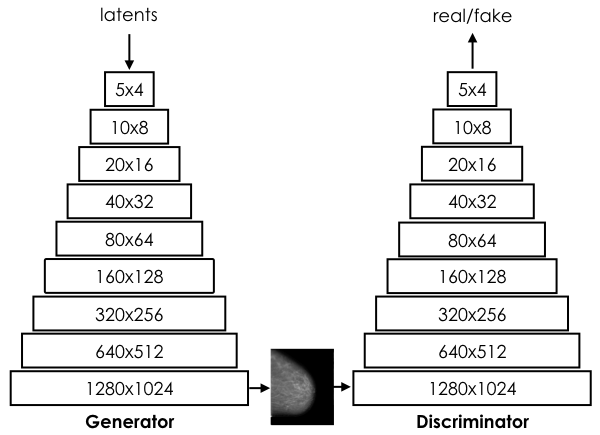}
    \caption{Scale 8}
  \end{subfigure}
  \caption{Illustration of the progressive growth of both networks during training.}
\end{figure}

\subsection{Progressive Training of GANs}

The research towards using GANs to synthesize ever increasing resolution of images has recently lead to a breakthrough in the work of \cite{pggan}. The underlying idea is to progressively increase the resolution of generated images by gradually adding new layers to the generator and discriminator networks. The generator first learns to synthesize the high-level structure and low frequency details of the image distribution, before gradually shifting its attention to finer details in higher scales. The fact that the generator does not need to learn all scales at once leads to increased stability. Progressive training also reduces training time, since most of the iterations are done at lower resolutions where the network sizes are small.

The original work includes several further important contributions. A dynamic weight initialization method is proposed to equalize the learning rate between parameters at different depths, batch normalization is substituted with a variant of local response normalization in order to constrain signal magnitudes in the generator, and a new evaluation metric is proposed (Sliced Wasserstein distance).

\subsection{Quantitative Evaluation Metrics}

There are two main factors we wish to assess in order to estimate the quality of outputs from the trained generator network. One is how probable the synthesized images are under the true data distribution, and the other is how large is the support of the generated distribution. Neither of these factors are straightforward to quantitatively assess and have been a subject of research since the advent of GANs.

The difficulty in assessing the fidelity to which the generated distribution follows the true data distribution stems from the fact:
\begin{itemize}
\item[--] We wish to compare sets of images, as opposed to pairs of images for which most image similarity metrics are designed.
\item[--] The comparison is based on conceptual attributes of appearance that are inherently subjective.
\end{itemize}

A first attempt to the problem was the consideration of the Inception Score (IS). Synthesized images $\boldsymbol{x}$ are presented to an ImageNet trained Inception model to produce a class prediction $y$, and a score is assigned based on the entropy of $p(y|\mathbf{x})$ and $p(y)$. Intuitively, high fidelity to the true distribution implies low entropy w.r.t. $p(y|\mathbf{x})$ (samples are unambiguous) and high distributional support translates to high entropy w.r.t. $p(y)$ (samples have high diversity).

An alternative to the IS, is the the Fr\`{e}chet Inception Distance (FID) \citep{FID}, which instead compares the distributions of the feature maps for original and generated images. The FID directly utilizes the training image dataset and can be more robust to transferring to images that were not used to train the inception model, e.g., facial images, as long as the features are also discriminative in the new domain.

An alternative metric, not requiring the use of a trained model, is the Multi-scale Structural Similarity Index (MS-SSIM) \citep{Odena2017, wang2003msssim}. The SSIM was designed to improve upon traditional image quality metrics and has been used as a loss function in deep learning, as it is differentiable \citep{MonoDepth}. In order to assess the quality of a trained GAN, it is necessary to randomly pair the original and generated images, compute the SSIM of each set and then compare with within set self-similarities.

Finally, an interesting alternative to the aforementioned metrics, proposed in \cite{pggan}, is the Multi-scale Sliced Wasserstein metric. The concept is to compare the sorted sets of descriptors extracted from original and generated images. In order to make this metric computationally efficient, the authors have used descriptors that correspond to random projections of image patches.

\section{Mammogram Synthesis} \label{proposed_approach}

\subsection{Clinical setting and data}

\begin{figure}[t!]
 \begin{center}
 \label{Figure 4}
 	\begin{subfigure}[t]{0.30\textwidth}
    	\includegraphics[width=1.0\textwidth]{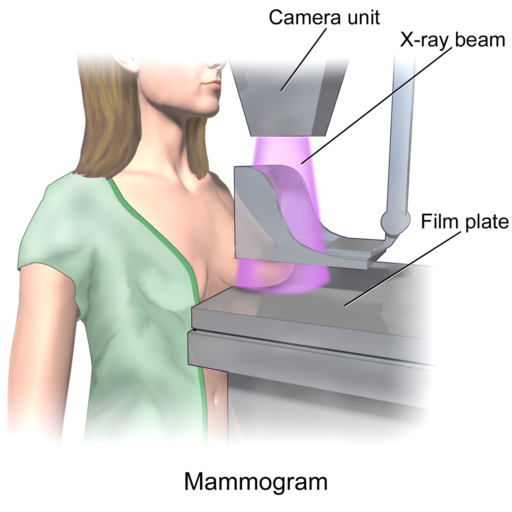}
    	\caption{Acquisition of CC image \citep{blausen2014}.}
    	\label{subfig:cc_illustration}
  	\end{subfigure}
    \begin{subfigure}[t]{0.50\textwidth}
    	\includegraphics[width=1.0\textwidth]{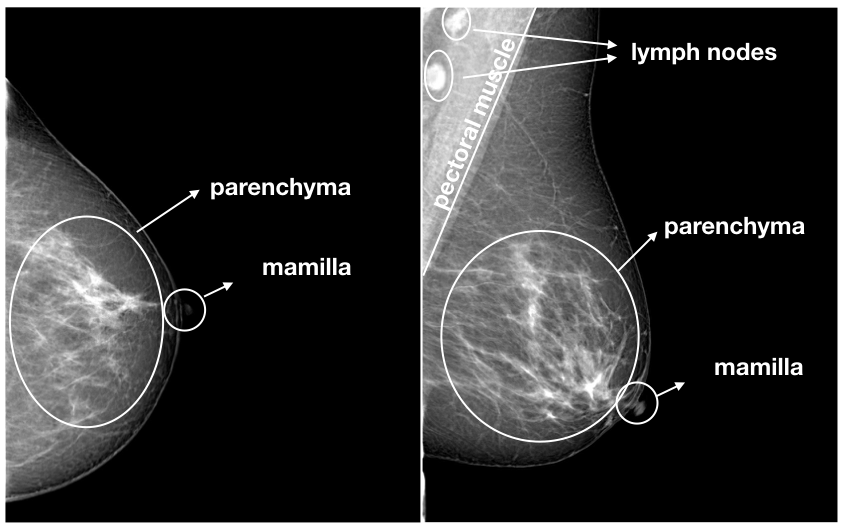}
    	\caption{CC and MLO view examples.}
    	\label{subfig:cc_and_mlo_example}
  	\end{subfigure}
 \end{center}
\end{figure}

Mammograms are relatively low-dose soft tissue X-rays of the breast. Acquisition is performed after each breast in turn has been flattened using two plastic paddles, as illustrated in Fig. \ref{subfig:cc_illustration}. Conventionally both left and right breasts are imaged using two standard views, the cranial-caudal (CC) and the mediolateral-oblique (MLO), which are shown in Fig. \ref{subfig:cc_and_mlo_example}. This results in a total of four 7-10 megapixel images per patient.

Hanging protocols are the series of actions performed to arrange images on a screen to be shown to the radiologist. Hanging protocols are designed to work across hardware and clinical sites. In mammography, this defines how to setup and present the images for the reader, including preferred windowing of image intensities and image size.

We have acquired a large number of images (>1,000,000) which we used for the purpose of this work. From this proprietary dataset we excluded images containing post-operative artifacts (metal clips, etc.) as well as large foreign bodies (pacemakers, implants, etc.). Otherwise, the images contain a wide variation in terms of anatomical differences and pathology (including benign and malignant cases) and the dataset corresponds to what is typically found in screening clinics.

\subsection{Training}

\begin{figure}[ht!]
\begin{center}
  \begin{subfigure}[t]{0.24\textwidth}
	\includegraphics[width=\textwidth]{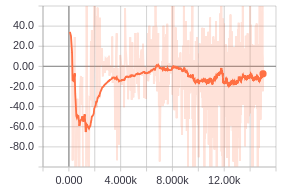}
    \caption{Discriminator binary cross entropy.}
  \end{subfigure}
  \begin{subfigure}[t]{0.24\textwidth}
	\includegraphics[width=\textwidth]{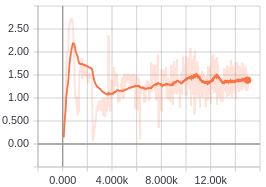}
    \caption{Gradient magnitudes contributing in Eq. (\ref{eq:wggp}).}
  \end{subfigure}
  \begin{subfigure}[t]{0.24\textwidth}
	\includegraphics[width=\textwidth]{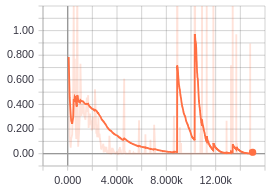}
    \caption{Label cross entropy for original images.}
  \end{subfigure}
  \begin{subfigure}[t]{0.24\textwidth}
	\includegraphics[width=\textwidth]{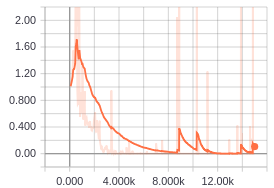}
    \caption{Label cross entropy for generated images.}
  \end{subfigure}
  \\
  \begin{subfigure}[b]{1.0\textwidth}
    \includegraphics[width=\textwidth]{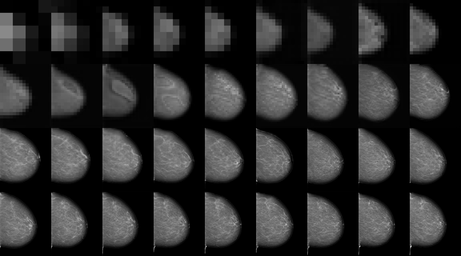}
    \caption{The training progression of a successful run.} \label{fig:training}
   \end{subfigure}
   \caption{Note that artifacts appear after around 4.7 million images have been presented to the network. Training recovers shortly after that, however, as can be seen in the diagnostic plots, this failure is not easily detectable from the curves.}
   \end{center}
\end{figure}
   
We used a simple preprocessing method that preserves both the original aspect ratio of each image and the hanging protocol. More specifically, we down-sampled by the largest factor to match one of the desired dimensions and padded the other dimension with zeros. The final image size is 1280x1024 pixels which (to the best of our knowledge) is the highest image resolution generated by a GAN thus far.

Despite using progressive training, we still had to overcome significant stability issues, due to the high resolution. We took several steps to maximize the probability of a successful run outlined in the following.

First, we increased the number of images used for training, from an initial 150k to 450k. This inevitably introduces more variation, along with some noise due to images that are erroneously included in the training set - some examples are shown in \ref{subfig:failures_real} of the Appendix. Nevertheless, we argue the extra information to be leveraged is beneficial for training.

Second, as suggested in \cite{Salimans2016}, we added some supervised information. More specifically we conditioned on the view, namely CC and MLO, which is highly relevant as it has significant impact on the visual appearance of the images.

Finally, we slightly decreased the learning rate from the one originally used in \cite{pggan}, from $0.002$ to $0.0015$ and gradually increased the discriminator iterations, from $1$ to a maximum of $5$ discriminator updates for each generator update. Even with these modifications, we had to often restart training and artifacts periodically appeared, but the network was able to recover in most cases. An example of the training progress for a successful run is shown in Fig. \ref{fig:training}. 

We performed our training on an NVIDIA DGX-1, with 8 V100 GPUs, 16GB GPU memory each. We initially trained until the network was presented with 15 million images, which is equivalent to 33 epochs which took about 52 hours. Then we resumed training for an additional 5 million images and selected the best network checkpoint based on the Sliced Wasserstein Distance \citep{pggan}.

\subsection{Results}

\begin{figure}[ht!]
  \begin{subfigure}[t]{1.0\textwidth}
    \includegraphics[width=1.0\textwidth]{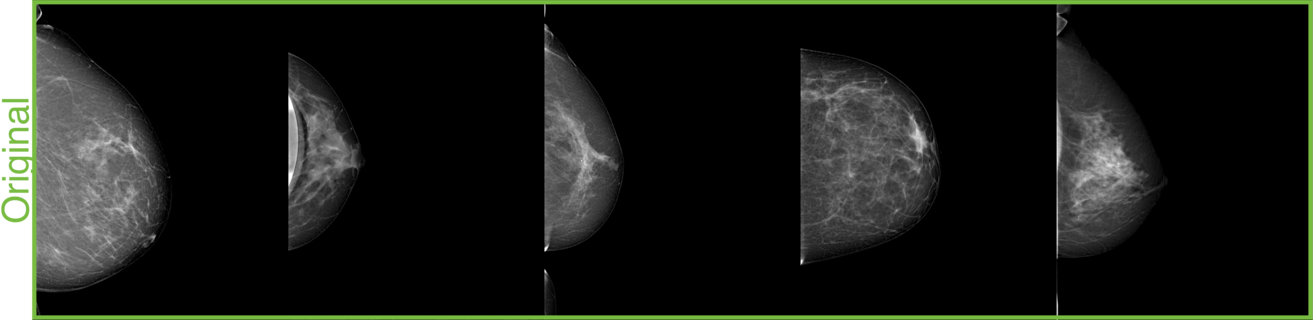}
    \includegraphics[width=1.0\textwidth]{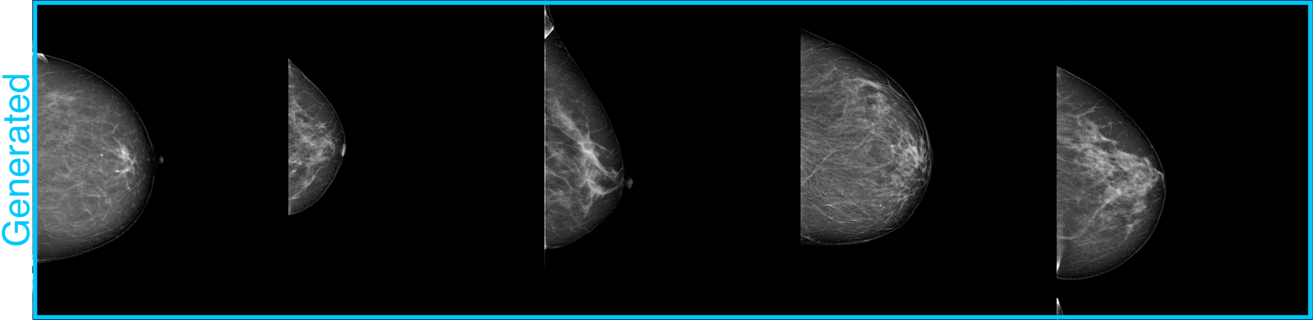}
    \caption{Randomly sampled examples of real and generated CC views.}
  \end{subfigure}
  \begin{subfigure}[b]{1.0\textwidth}
  	\includegraphics[width=1.0\textwidth]{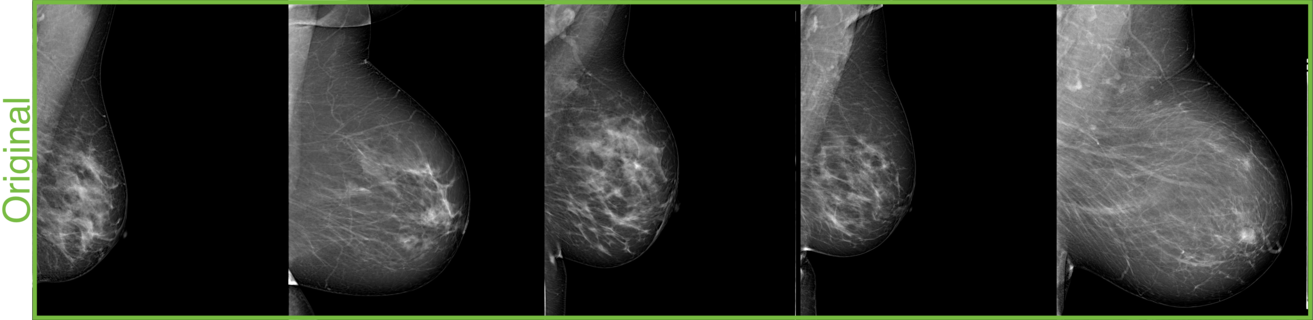}
    \includegraphics[width=1.0\textwidth]{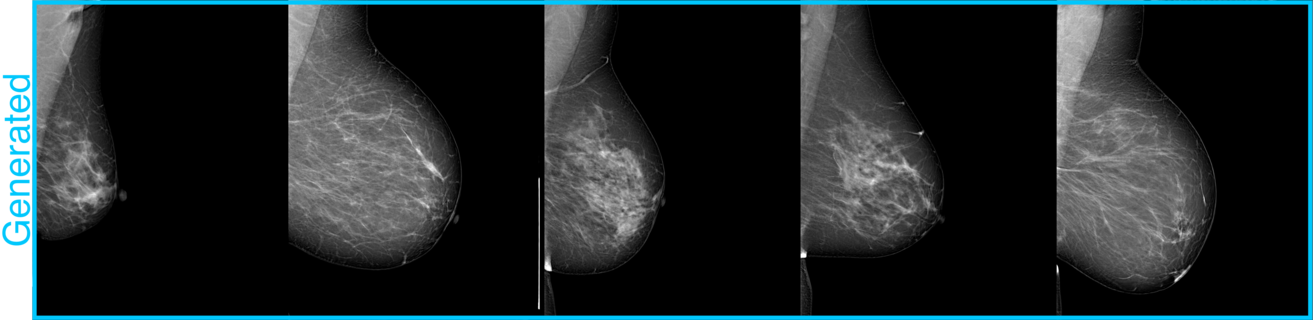}
    \caption{Randomly sampled examples of real and generated MLO views.}
  \end{subfigure}
  \caption{Examples of generated images from the GAN.}
\end{figure}

The final samples drawn from a successfully trained network look very promising. Most of the generated images seem highly realistic with a broad range of inter-image variability, which indicates good representation of the underlying true distribution. However, we also observed some common artifacts and failures, which we discuss below.

For more visual examples we refer to the Appendix, where we present images in several different format, described as follows:
\begin{itemize}
	\item[--] 6x5 grids of randomly selected generations from CC and MLO views (Fig. \ref{fig:cc_random_generated_6x5} and \ref{fig:mlo_random_generated_6x5}).
    \item[--] 5x2 grids of randomly selected generations from CC and MLO views, alongside randomly selected real images. In this case, we also indicatively mark the best and worst generations (Fig. \ref{fig:cc_random_generated_5x2} and \ref{fig:mlo_random_generated_5x2}).
    \item[--] 3x5 grids of handpicked convincing results from CC and MLO views (Fig. \ref{fig:handpicked_generated_3x5}).
        \item[--] 1x3 grids of handpicked convincing results, alongside real images (Fig. \ref{fig:cc_handpicked_3x1} and \ref{fig:mlo_handpicked_3x1}).
    \item[--] 2x5 grids where we present examples of failures from CC and MLO views, along with images with artifacts from the training set (Fig. \ref{fig:failures}).
\end{itemize}

We also refer to a video illustrating a random walk through the latent space that can be found here: \url{https://www.youtube.com/watch?v=Ro-tZ6wYn1g}

\begin{figure}[t!]
  \begin{center}
      \includegraphics[width=0.75\textwidth]{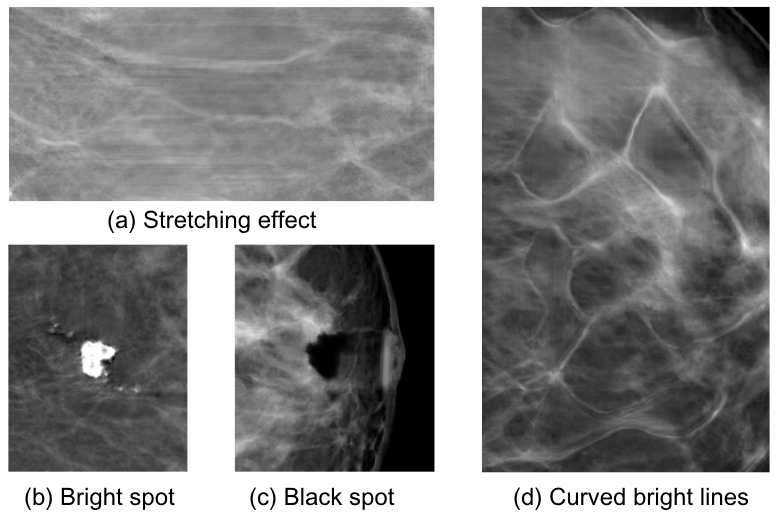}
      \caption{Most commonly seen artifact patterns}
  \end{center}
\end{figure}

\begin{figure}[b!]
  \begin{subfigure}[t]{0.3\textwidth}
    \includegraphics[width=\textwidth]{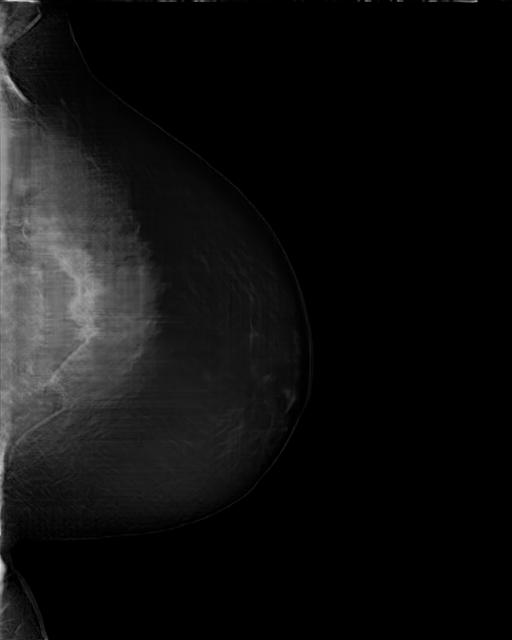}
    \caption{Transitioning towards a larger size.}
  \end{subfigure}
  \rulesep
  \begin{subfigure}[t]{0.3\textwidth}
    \includegraphics[width=\textwidth]{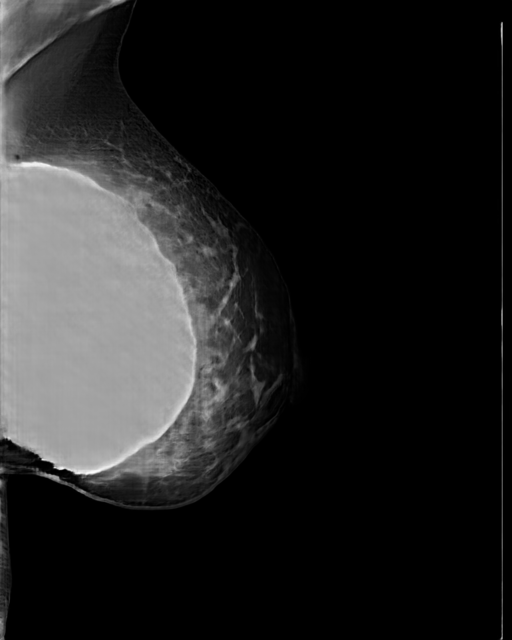}
    \caption{Attempted reproduction of breast implant.}
  \end{subfigure}
  \rulesep
  \begin{subfigure}[t]{0.3\textwidth}
    \includegraphics[width=\textwidth]{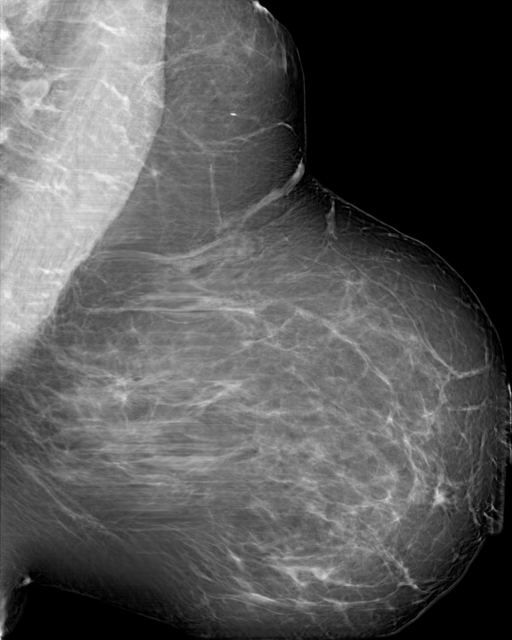}
    \caption{Distorted reproduction near the right hand side border of the image.}
  \end{subfigure}
  \caption{Common failures.}
\end{figure}

\paragraph{Views}

The MLO view is evidently the harder one to model, unsurprisingly so, as it exhibits the highest variation and contains the most anatomical information, with the pectoral muscle clearly visible, lymph nodes in some cases, and of course the breast parenchyma (Fig. \ref{subfig:cc_and_mlo_example})

Samples from the CC view seem subjectively of higher quality, due to their relative simplicity compared the MLO view.

\paragraph{Calcifications and metal markers}

Calcifications are caused naturally in the breast from calcium deposition and can vary in size and shape, but appear very bright (white) on the image as they fully absorb passing X-rays. They are important in mammography as certain patterns can be a strong indication of malignancy, while others are benign (e.g., vascular deposits).

\begin{wrapfigure}{h!}{0.3\textwidth}
\begin{center}
  \vspace{-10pt}
  \includegraphics[width=0.25\textwidth]{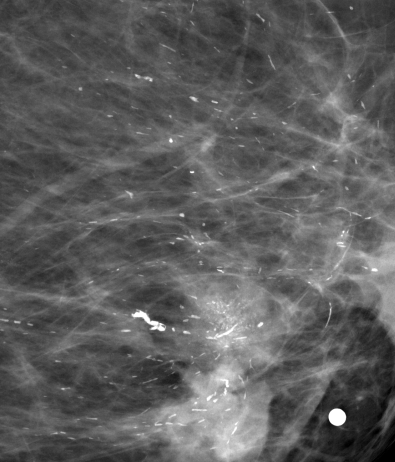}
  \caption{Calcifications and a round marker (bottom right) commonly used by the technician to indicate a palpable lesion.}
  \vspace{-40pt}
  \label{fig:calcs_and_marker}
 \end{center}
\end{wrapfigure}

External skin markers are frequently used by technicians performing the mammogram to indicate the position of a palpable lesion in the breast for the attention of the radiologists who is going to perform the reading. They also appear very bright, but are distinctively fully circular in shape.

In Fig. \ref{fig:calcs_and_marker} we show an example of both calcifications and a marker in the bottom right, appearing in the same image.

We have observed that the generator strongly resists these structures. It is only in very late stages of training that features roughly similar to medium sized calcification may appear in the generations, but they are not very convincing. We assume that the network architecture acts as a strong prior against such features, which do not appear in natural images (as also suggested in \cite{Ulyanov_2018_CVPR}).

\paragraph{Common artifacts}

We observe several types of failures in the generated images. Some of them are clearly network failures, which indicate that not all possible latent vectors correspond to valid images in pixel space. Others can be attributed to problems in the training set. Examples of such images are shown in Fig. \ref{subfig:failures_real}.

\section{Conclusion}

In this work we present our methodology for generating highly realistic, high-resolution synthetic mammograms using a progressively trained generative adversarial network (GAN). Generative models can be especially valuable for medical imaging research. However, GANs have not so far been able to scale to the high resolution required in FFDM. We have managed to overcome the underlying instabilities inherent in training such adversarial models and have been able to generate images of highest resolution reported so far, namely 1280x1024 pixels. We have identified a number of limitations, including common artifacts and failure cases, indicating that further research is required but that promising results can already be achieved. We hope this work can serve as a useful guide and facilitate further research on GANs in the medical imaging domain.

\subsubsection*{Acknowledgments}

The authors would like to thank Dr. Andreas Heindl and Dr. Galvin Khara for their valuable inputs and insights, Nicolas Markos for his valuable expertise in high performance computing, as well as the rest of their colleagues at Kheiron Medical Technologies Ltd. for their help and support.

\bibliographystyle{hapalike}
\bibliography{bibliography.bib}

\begin{thebibliography}{}

\bibitem[Adolphs et~al., 2018]{Adolphs2018}
Adolphs, L., Daneshmand, H., Lucchi, A., and Hofmann, T. (2018).
\newblock {Local Saddle Point Optimization: A Curvature Exploitation Approach}.
\newblock (2).

\bibitem[Arjovsky et~al., 2017]{WassersteinGAN}
Arjovsky, M., Chintala, S., and Bottou, L. (2017).
\newblock {Wasserstein GAN}.
\newblock {\em arXiv preprint arXiv:1701.07875 (2017)}.

\bibitem[Blausen, 2014]{blausen2014}
Blausen, M. (2014).
\newblock Blausen gallery 2014.
\newblock {\em Wikiversity Journal of Medicine}, 1(2).

\bibitem[Costa et~al., 2017]{Costa2018}
Costa, P., Galdran, A., Meyer, M.~I., Niemeijer, M., Abr{\`a}moff, M.,
  Mendon{\c{c}}a, A.~M., and Campilho, A. (2017).
\newblock End-to-end adversarial retinal image synthesis.
\newblock {\em IEEE transactions on medical imaging}, 37(3):781--791.

\bibitem[Daskalakis et~al., 2018]{daskalakis2018}
Daskalakis, C., Ilyas, A., Syrgkanis, V., and Zeng, H. (2018).
\newblock {Training GANs with Optimism}.
\newblock In {\em Proceedings of the International Conference on Learning
  Representations}.

\bibitem[Elgammal et~al., 2017]{elgammal2017}
Elgammal, A., Liu, B., Elhoseiny, M., and Mazzone, M. (2017).
\newblock {CAN: Creative Adversarial Networks, Generating "Art" by Learning
  About Styles and Deviating from Style Norms}.
\newblock {\em arXiv preprint arXiv:1706.07068 (2017)}.

\bibitem[Frid-Adar et~al., 2018]{frid2018synthetic}
Frid-Adar, M., Klang, E., Amitai, M., Goldberger, J., and Greenspan, H. (2018).
\newblock Synthetic data augmentation using gan for improved liver lesion
  classification.
\newblock In {\em IEEE 15th International Symposium on Biomedical Imaging (ISBI
  2018)}, pages 289--293.

\bibitem[Godard et~al., 2017]{MonoDepth}
Godard, C., Mac~Aodha, O., and Brostow, G.~J. (2017).
\newblock {Unsupervised monocular depth estimation with left-right
  consistency}.
\newblock pages 6602--6611.

\bibitem[Goodfellow et~al., 2014]{goodfellow2014}
Goodfellow, I., Pouget-Abadie, J., Mirza, M., Xu, B., Warde-Farley, D., Ozair,
  S., Courville, A., and Bengio, Y. (2014).
\newblock Generative adversarial nets.
\newblock In {\em Advances in neural information processing systems}, pages
  2672--2680.

\bibitem[Gulrajani et~al., 2017]{ImprovedWassersteinGAN}
Gulrajani, I., Ahmed, F., Arjovsky, M., Dumoulin, V., and Courville, A.~C.
  (2017).
\newblock Improved training of wasserstein gans.
\newblock In {\em Advances in Neural Information Processing Systems}, pages
  5767--5777.

\bibitem[Heusel et~al., 2017]{FID}
Heusel, M., Ramsauer, H., Unterthiner, T., Nessler, B., and Hochreiter, S.
  (2017).
\newblock Gans trained by a two time-scale update rule converge to a local nash
  equilibrium.
\newblock In {\em Advances in Neural Information Processing Systems}, pages
  6626--6637.

\bibitem[Isola et~al., 2017]{isola2016}
Isola, P., Zhu, J.~Y., Zhou, T., and Efros, A.~A. (2017).
\newblock {Image-to-image translation with conditional adversarial networks}.
\newblock pages 5967--5976.

\bibitem[Kamnitsas et~al., 2017]{kamnitsas2017unsupervised}
Kamnitsas, K., Baumgartner, C., Ledig, C., Newcombe, V., Simpson, J., Kane, A.,
  Menon, D., Nori, A., Criminisi, A., Rueckert, D., et~al. (2017).
\newblock Unsupervised domain adaptation in brain lesion segmentation with
  adversarial networks.
\newblock In {\em International Conference on Information Processing in Medical
  Imaging}, pages 597--609. Springer.

\bibitem[Karras et~al., 2018]{pggan}
Karras, T., Aila, T., Laine, S., and Lehtinen, J. (2018).
\newblock Progressive growing of gans for improved quality, stability, and
  variation.
\newblock In {\em Proceedings of the International Conference on Learning
  Representations}.

\bibitem[Kingma and Welling, 2013]{kingma2013}
Kingma, D.~P. and Welling, M. (2013).
\newblock {Auto-Encoding Variational Bayes}.
\newblock {\em arXiv preprint arXiv:1312.6114 (2013)}.

\bibitem[Lafarge et~al., 2017]{lafarge2017domain}
Lafarge, M.~W., Pluim, J.~P., Eppenhof, K.~A., Moeskops, P., and Veta, M.
  (2017).
\newblock Domain-adversarial neural networks to address the appearance
  variability of histopathology images.
\newblock In {\em Deep Learning in Medical Image Analysis and Multimodal
  Learning for Clinical Decision Support}, pages 83--91. Springer.

\bibitem[Lahiri et~al., 2017]{lahiri2017generative}
Lahiri, A., Ayush, K., Biswas, P.~K., and Mitra, P. (2017).
\newblock Generative adversarial learning for reducing manual annotation in
  semantic segmentation on large scale miscroscopy images: Automated vessel
  segmentation in retinal fundus image as test case.
\newblock In {\em IEEE Computer Society Conference on Computer Vision and
  Pattern Recognition Workshops}, pages 42--48.

\bibitem[Ledig et~al., 2017]{ledig2017superres}
Ledig, C., Theis, L., Huszar, F., Caballero, J., Cunningham, A., Acosta, A.,
  Aitken, A., Tejani, A., Totz, J., Wang, Z., et~al. (2017).
\newblock Photo-realistic single image super-resolution using a generative
  adversarial network.
\newblock In {\em Proceedings of the IEEE Conference on Computer Vision and
  Pattern Recognition}, pages 4681--4690.

\bibitem[Odena et~al., 2017]{Odena2017}
Odena, A., Olah, C., and Shlens, J. (2017).
\newblock Conditional image synthesis with auxiliary classifier gans.
\newblock In {\em International Conference on Machine Learning}, pages
  2642--2651.

\bibitem[Salehinejad et~al., 2018]{xrayGAN}
Salehinejad, H., Valaee, S., Dowdell, T., Colak, E., and Barfett, J. (2018).
\newblock {Generalization of Deep Neural Networks for Chest Pathology
  Classification in X-Rays Using Generative Adversarial Networks}.
\newblock In {\em IEEE International Conference on Acoustics, Speech and Signal
  Processing (ICASSP)}.

\bibitem[Salimans et~al., 2016]{Salimans2016}
Salimans, T., Goodfellow, I., Zaremba, W., Cheung, V., Radford, A., and Chen,
  X. (2016).
\newblock Improved techniques for training gans.
\newblock In {\em Advances in Neural Information Processing Systems}, pages
  2234--2242.

\bibitem[Ulyanov et~al., 2017]{ulyanov2017texture}
Ulyanov, D., Vedaldi, A., and Lempitsky, V. (2017).
\newblock Improved texture networks: Maximizing quality and diversity in
  feed-forward stylization and texture synthesis.
\newblock In {\em Proceedings of the IEEE Conference on Computer Vision and
  Pattern Recognition}, volume~1, page~6.

\bibitem[Ulyanov et~al., 2018]{Ulyanov_2018_CVPR}
Ulyanov, D., Vedaldi, A., and Lempitsky, V. (2018).
\newblock Deep image prior.
\newblock In {\em The IEEE Conference on Computer Vision and Pattern
  Recognition (CVPR)}.

\bibitem[Van~Oord et~al., 2016]{van2016pixel}
Van~Oord, A., Kalchbrenner, N., and Kavukcuoglu, K. (2016).
\newblock Pixel recurrent neural networks.
\newblock In {\em International Conference on Machine Learning}, pages
  1747--1756.

\bibitem[Villani, 2009]{OptTrans}
Villani, C. (2009).
\newblock {\em {Optimal transport : old and new}}.
\newblock Springer.

\bibitem[Wang et~al., 2003]{wang2003msssim}
Wang, Z., Simoncelli, E., Bovik, A., et~al. (2003).
\newblock Multi-scale structural similarity for image quality assessment.
\newblock In {\em Asilomar Conference on Signals, Systems, and Computers},
  volume~2, pages 1398--1402.

\bibitem[Wolterink et~al., 2017]{wolterink2017generative}
Wolterink, J.~M., Leiner, T., Viergever, M.~A., and I{\v{s}}gum, I. (2017).
\newblock Generative adversarial networks for noise reduction in low-dose ct.
\newblock {\em IEEE transactions on medical imaging}, 36(12):2536--2545.

\bibitem[Yadav et~al., 2018]{yadav2018stabilizing}
Yadav, A., Shah, S., Xu, Z., Jacobs, D., and Goldstein, T. (2018).
\newblock Stabilizing adversarial nets with prediction methods.
\newblock In {\em Proceedings of the International Conference on Learning
  Representations}.

\bibitem[Yi and Babyn, 2018]{low_dose_CT_2}
Yi, X. and Babyn, P. (2018).
\newblock Sharpness-aware low-dose ct denoising using conditional generative
  adversarial network.
\newblock {\em Journal of digital imaging}, pages 1--15.

\end{thebibliography}

\newpage

\begin{appendices}
\appendix
\section{Further examples}
\begin{figure}[h]
  \begin{center}
    \includegraphics[width=0.92\textwidth]{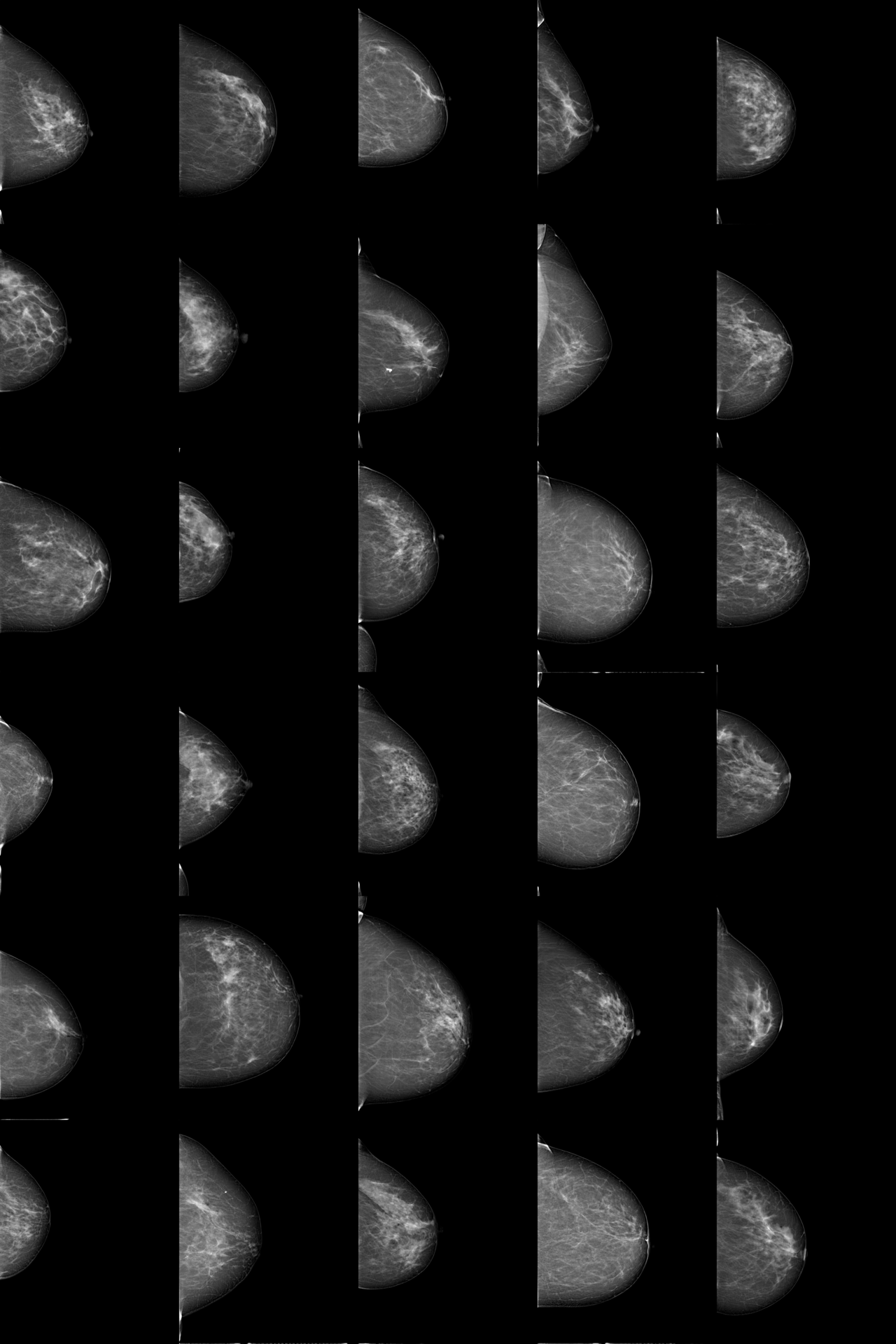}
    \caption{Random samples of generated CC views.}
    \label{fig:cc_random_generated_6x5}
    \end{center}
\end{figure}

\begin{figure}[h]
  \begin{center}
    \includegraphics[width=0.92\textwidth]{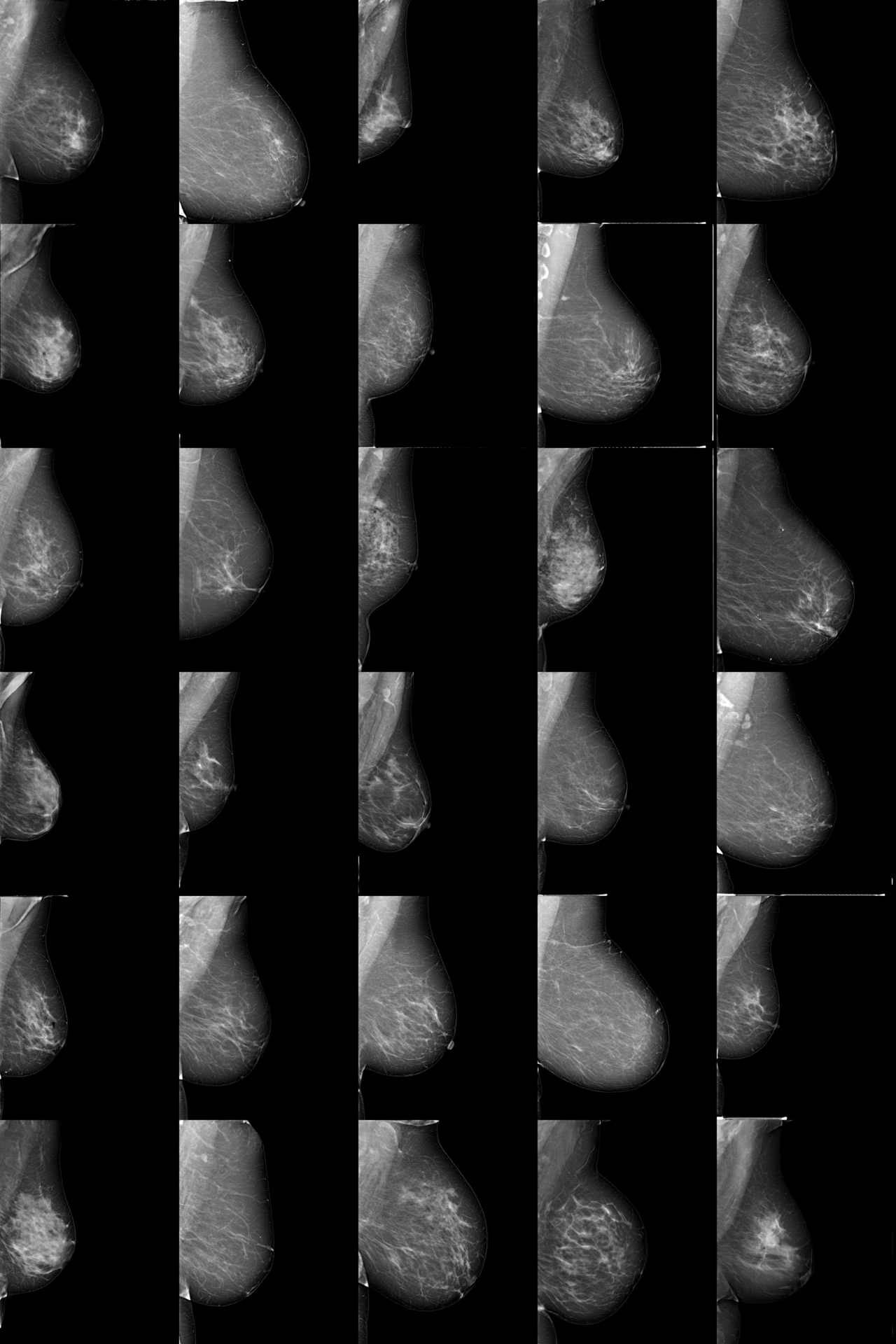}
    \caption{Random samples of generated MLO views.}
    \label{fig:mlo_random_generated_6x5}
    \end{center}
\end{figure}

\begin{figure}[h]
  \begin{subfigure}[t]{0.48\textwidth}
    \includegraphics[width=\textwidth]{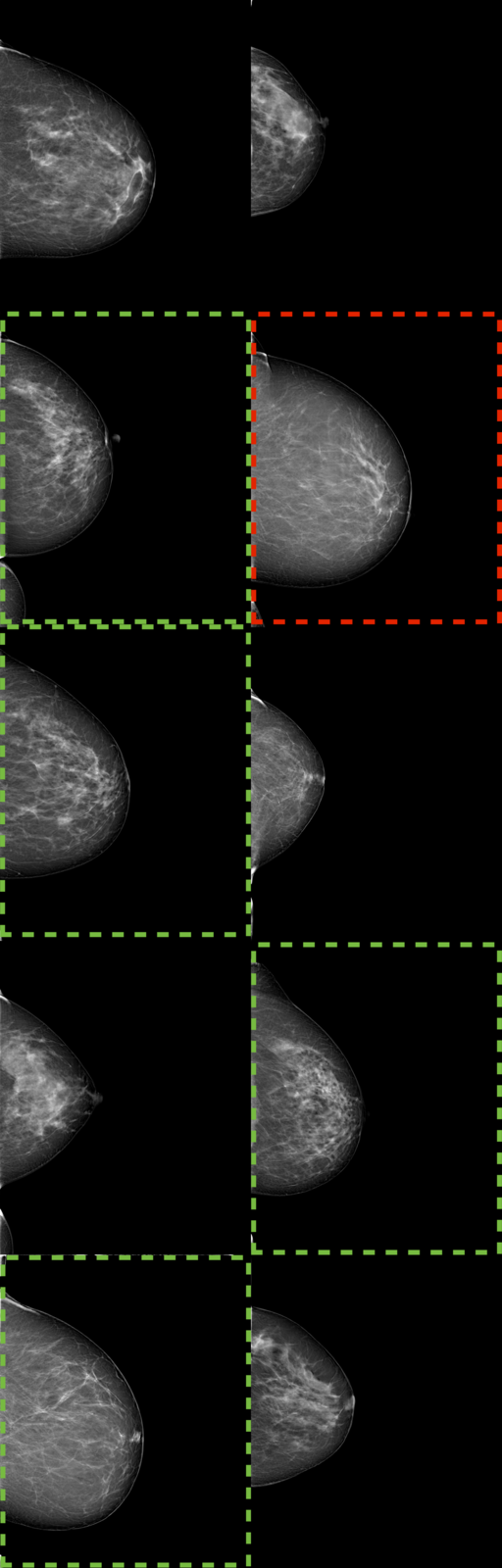}
    \caption{Generated}
  \end{subfigure}
  \rulesep
  \begin{subfigure}[t]{0.48\textwidth}
    \includegraphics[width=\textwidth]{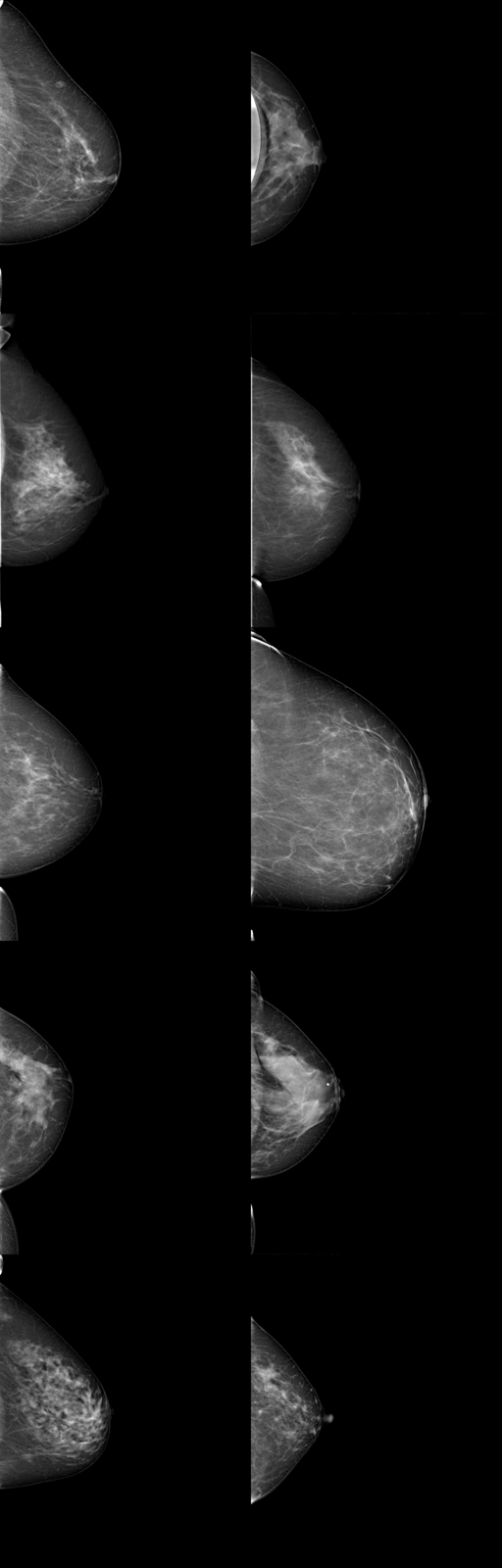}
    \caption{Original}
  \end{subfigure}
  \caption{Randomly sampled original and generated CC views. The green dashed line denotes particularly convincing samples and the red dashed line denotes images with obvious artifacts.} 
  \label{fig:cc_random_generated_5x2}
\end{figure}

\begin{figure}[h]
  \begin{subfigure}[t]{0.48\textwidth}
    \includegraphics[width=\textwidth]{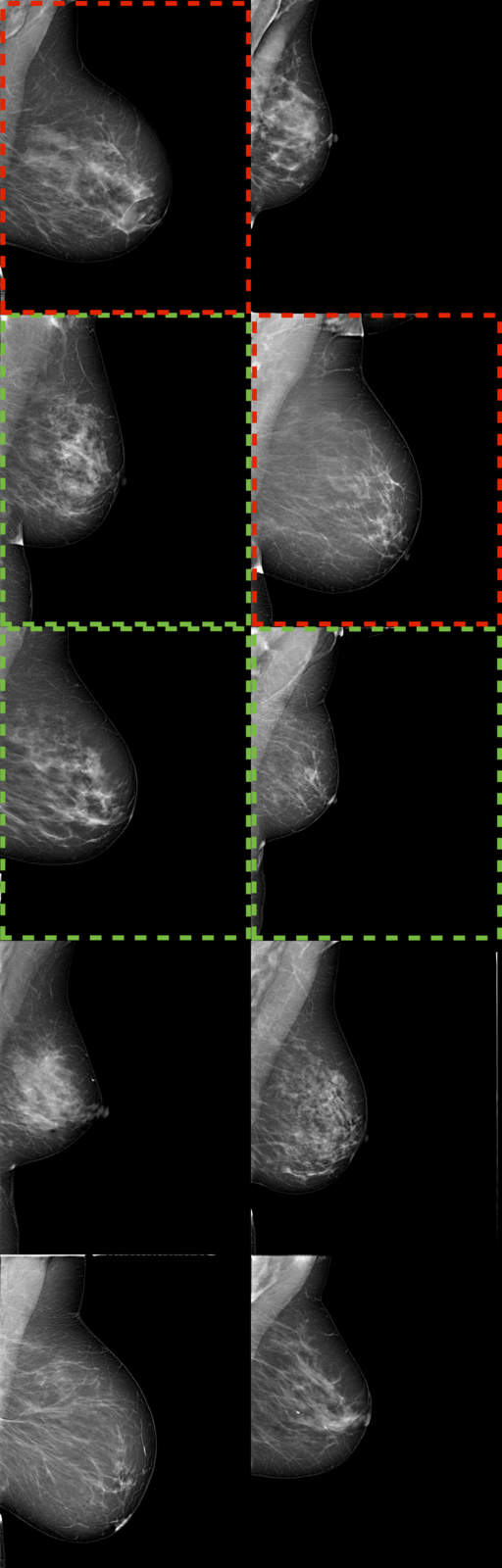}
    \caption{Generated}
  \end{subfigure}
  \rulesep
  \begin{subfigure}[t]{0.48\textwidth}
    \includegraphics[width=\textwidth]{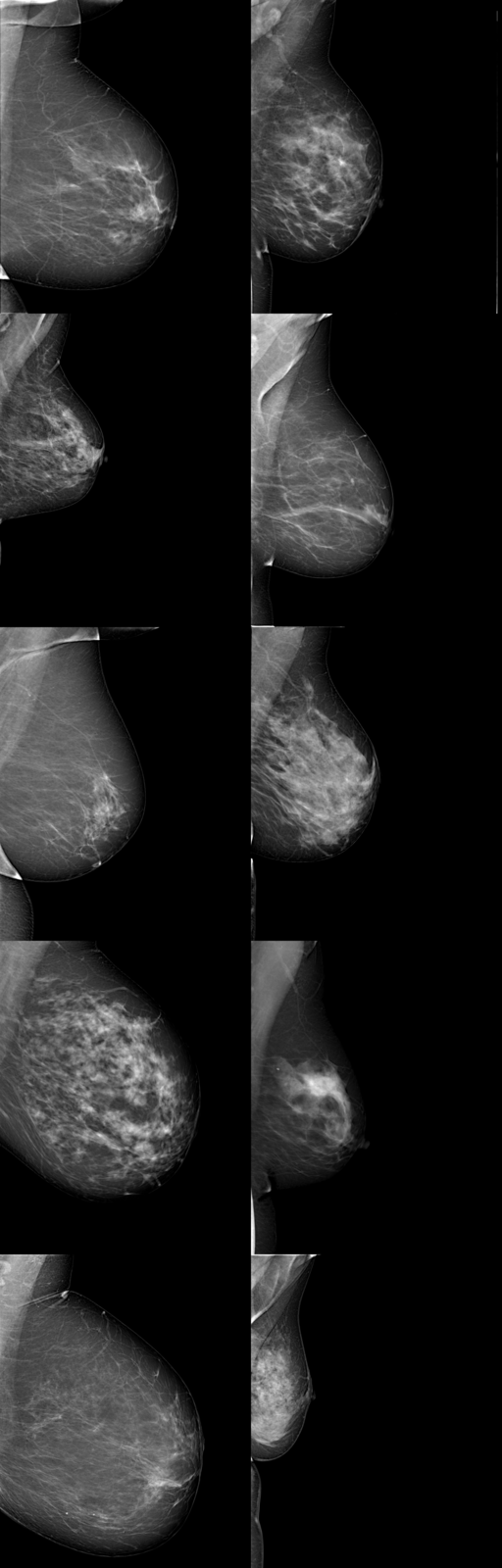}
    \caption{Original}
  \end{subfigure}
  \caption{Randomly sampled original and generated MLO views. The green dashed line denotes particularly convincing samples and the red dashed line denotes images with obvious artifacts.}
  \label{fig:mlo_random_generated_5x2}
\end{figure}

\begin{figure}[h]
  \begin{subfigure}[t]{1.00\textwidth}
    \includegraphics[width=1.0\textwidth]{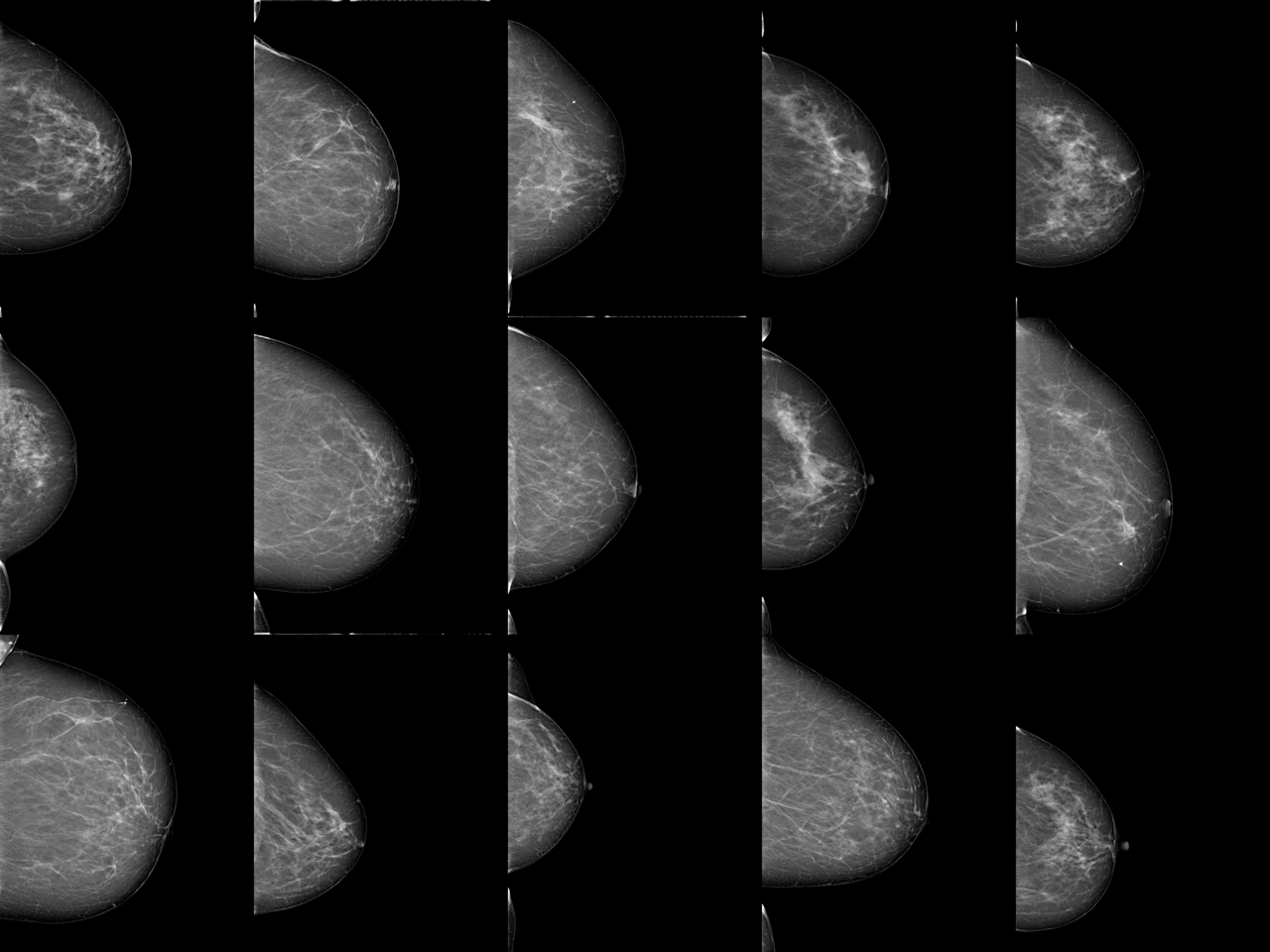}
    \caption{Hand-picked examples of generated CC views.}
  \end{subfigure}
  \\
  \begin{subfigure}[t]{1.00\textwidth}
    \includegraphics[width=1.0\textwidth]{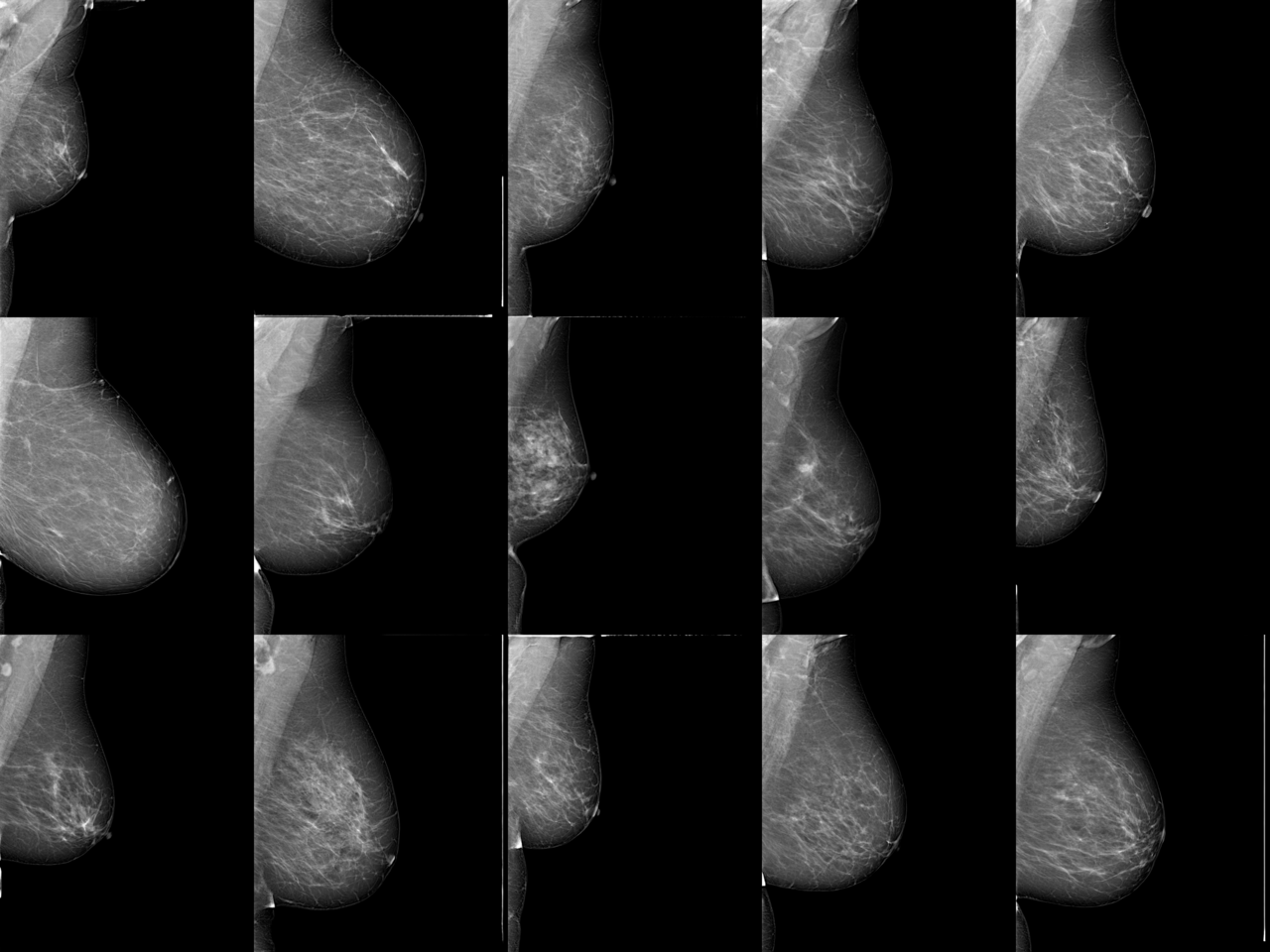}
    \caption{Generated images from MLO view.}
  \end{subfigure}
  \caption{Handpicked examples of both CC and MLO views.}
  \label{fig:handpicked_generated_3x5}
\end{figure}

\begin{figure}[h]
 \begin{center}
  \begin{subfigure}[t]{0.42\textwidth}
    \includegraphics[width=\textwidth]{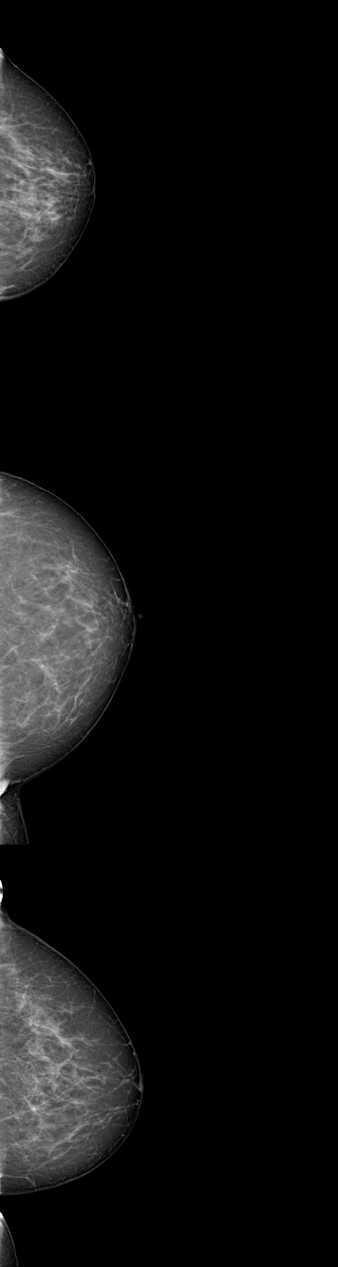}
    \caption{Generated}
  \end{subfigure}
  \rulesep
  \begin{subfigure}[t]{0.42\textwidth}
    \includegraphics[width=\textwidth]{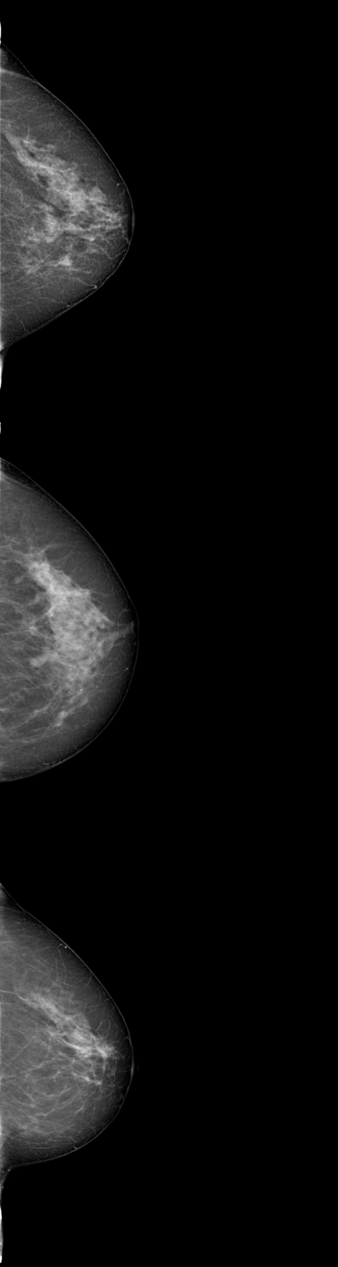}
    \caption{Original}
  \end{subfigure}
  \caption{Handpicked generated CC views alongside random original CC views.}
  \label{fig:cc_handpicked_3x1}
 \end{center}
\end{figure}

\begin{figure}[h]
 \begin{center}
  \begin{subfigure}[t]{0.42\textwidth}
    \includegraphics[width=\textwidth]{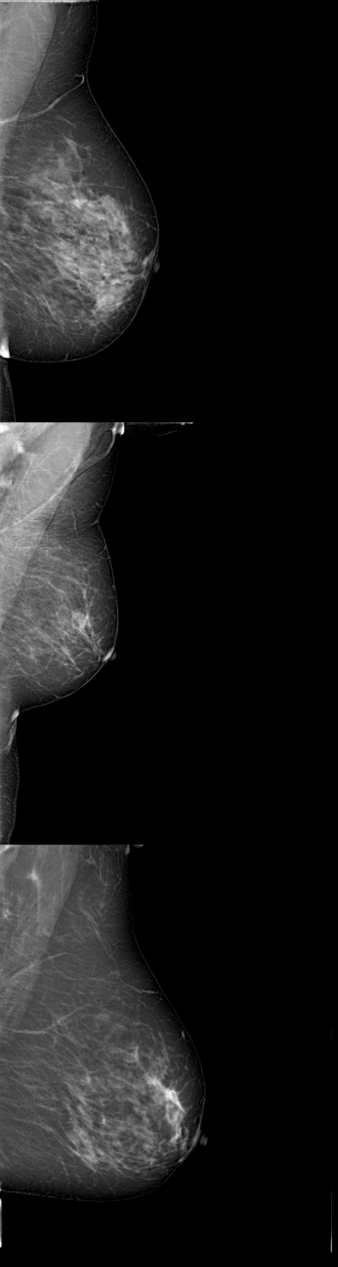}
    \caption{Generated}
  \end{subfigure}
  \rulesep
  \begin{subfigure}[t]{0.42\textwidth}
    \includegraphics[width=\textwidth]{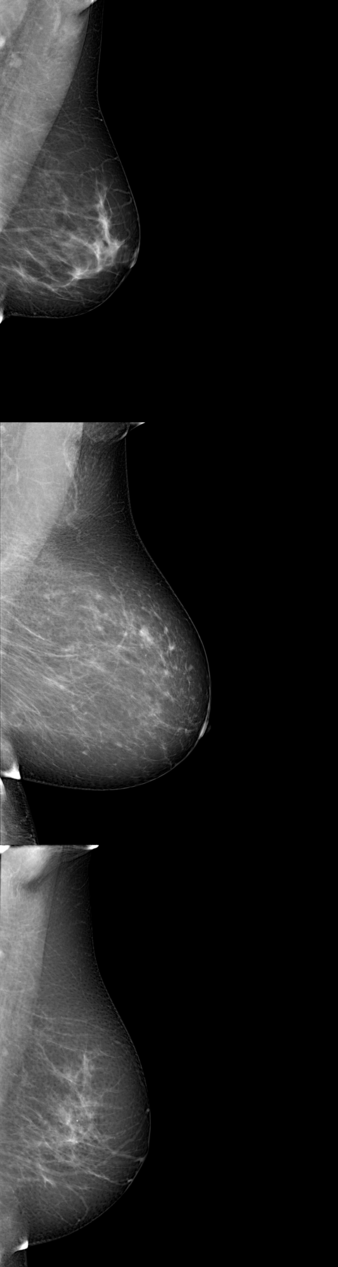}
    \caption{Original}
  \end{subfigure}
  \caption{Handpicked generated CC views alongside random original CC views.}
  \label{fig:mlo_handpicked_3x1}
 \end{center}
\end{figure}

\begin{figure}[h]
  \begin{subfigure}[t]{1.00\textwidth}
    \includegraphics[width=1.0\textwidth]{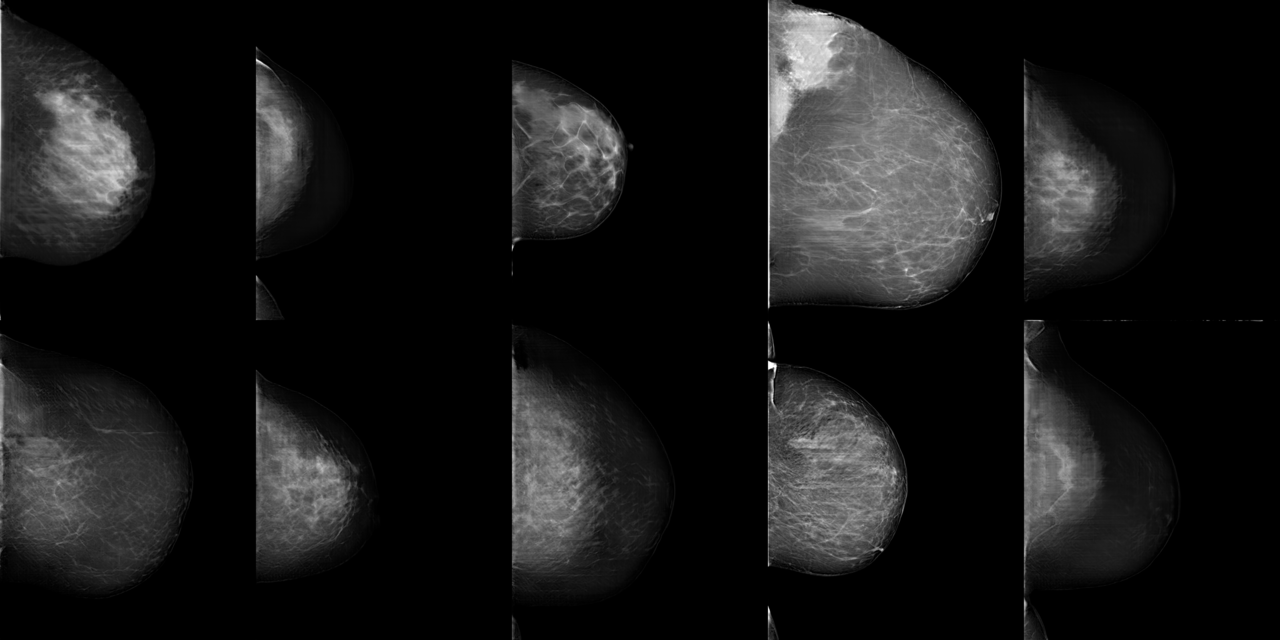}
    \caption{Worst examples of generated CC views.}
  \end{subfigure}
  \\
  \begin{subfigure}[t]{1.00\textwidth}
    \includegraphics[width=1.0\textwidth]{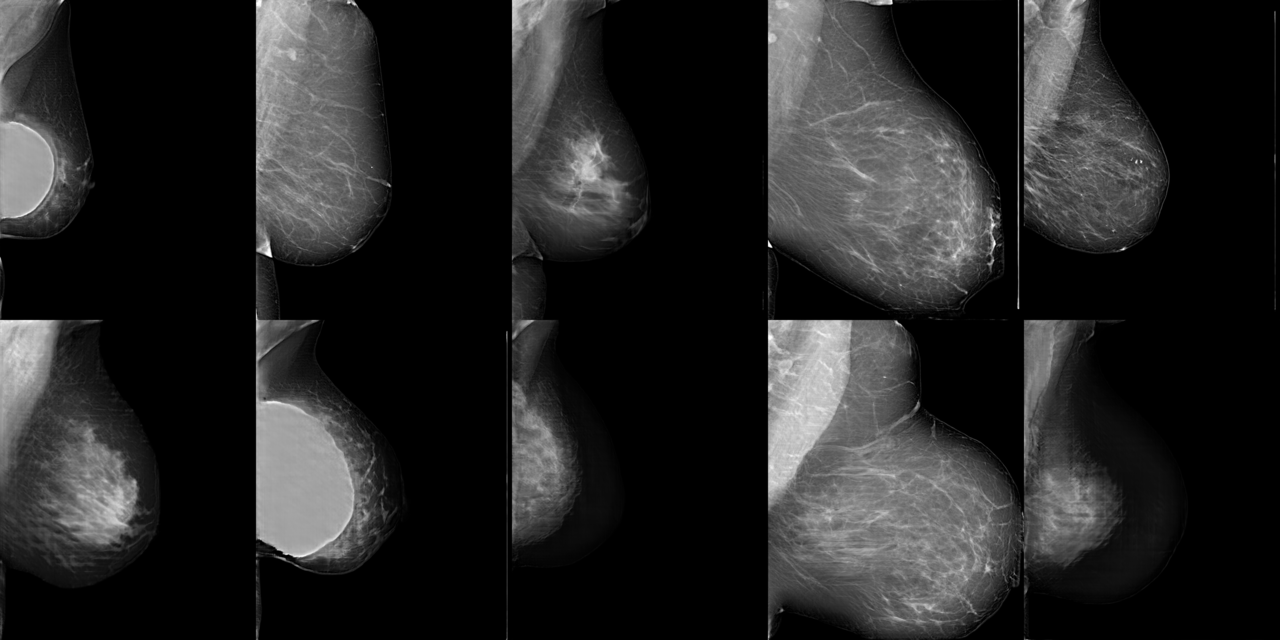}
    \caption{Worst examples of generated MLO views.}
  \end{subfigure}
  \\
  \begin{subfigure}[t]{1.00\textwidth}
    \includegraphics[width=1.0\textwidth]{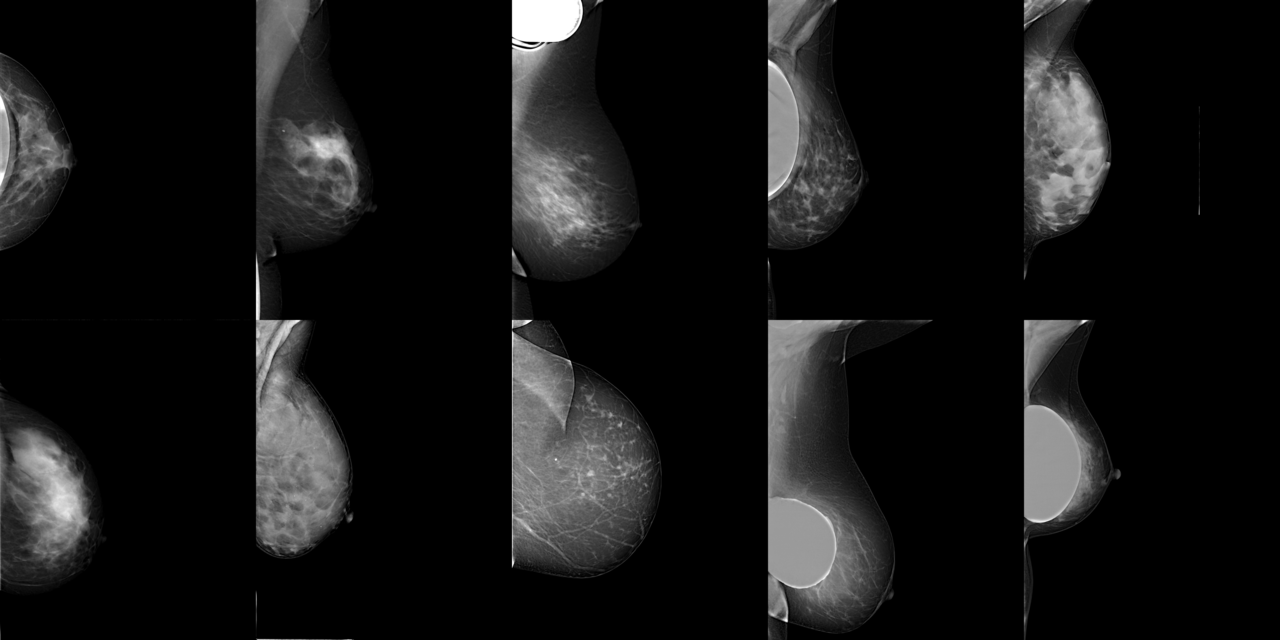}
    \caption{Original images with problematic appearances.}
    \label{subfig:failures_real}
  \end{subfigure}
  \caption{Worst examples we could find from both CC and MLO views.}
  \label{fig:failures}
\end{figure}

\end{appendices}

\end{document}